\documentclass[11pt]{article}

\usepackage[preprint]{acl}

\usepackage{times}
\usepackage{latexsym}

\usepackage[T1]{fontenc}
\usepackage[utf8]{inputenc}

\usepackage{microtype}

\usepackage{inconsolata}

\usepackage{graphicx}

\usepackage{amsmath,amsfonts}
\usepackage{multirow}
\usepackage{adjustbox}
\usepackage{amssymb}
\usepackage[breakable]{tcolorbox}
\usepackage{booktabs}
\usepackage[table]{xcolor}
\usepackage{pifont}
\newcommand{\cmark}{\ding{51}}  % 对勾
\newcommand{\xmark}{\ding{55}}  % 叉号
\title{Evolving and Executing Research Plans via Double-Loop Multi-Agent Collaboration}

\author{
  \textbf{Zhi Zhang\textsuperscript{1}},
  \textbf{Yan Liu\textsuperscript{1}}\thanks{Corresponding author.},
  \textbf{Zhejing Hu\textsuperscript{1}},
  \textbf{Gong Chen\textsuperscript{2}},
  \textbf{Sheng-hua Zhong\textsuperscript{3}},
  \textbf{Jiannong Cao\textsuperscript{1}}
\\
  \textsuperscript{1}Department of Computing, The Hong Kong Polytechnic University, Hong Kong, China\\
  \textsuperscript{2}FireTorch Partners, Shenzhen, China\\
  \textsuperscript{3}College of Computer Science and Software Engineering, Shenzhen University, Shenzhen, China
\\
  \small
    \texttt{\{zhi271.zhang, zhejing.hu\}@connect.polyu.hk}, 
    \texttt{\{yan.liu, jiannong.cao\}@polyu.edu.hk},\\
  \small
    \texttt{heinz@clozzz.com}, 
    \texttt{csshzhong@szu.edu.cn}
}

\begin{document}
\maketitle
\begin{abstract}
Automating the end-to-end scientific research process poses a fundamental challenge: it requires both evolving high-level plans that are novel and sound, and executing these plans correctly amidst dynamic and uncertain conditions. To address this bilevel challenge, we propose a novel Double-Loop Multi-Agent (DLMA) framework to solve the given research problem automatically. The leader loop, composed of professor agents, is responsible for evolving research plans. It employs an evolutionary algorithm through involvement, improvement, and integration meetings to iteratively generate and refine a pool of research proposals, exploring the solution space effectively. The follower loop, composed of doctoral student agents, is responsible for executing the best-evolved plan. It dynamically adjusts the plan during implementation via pre-hoc and post-hoc meetings, ensuring each step (e.g., drafting, coding) is well-supported by contextual and external observations. Extensive experiments on benchmarks like ACLAward and Laboratory show that DLMA generates research papers that achieve state-of-the-art scores in automated evaluation, significantly outperforming strong baselines. Ablation studies confirm the critical roles of both loops, with evolution driving novelty and execution ensuring soundness.
\end{abstract}

\section{Introduction}

Scientific research drives innovation, advances knowledge, and improves human lives \cite{wang2023scientific}. However, the research process remains labor-intensive, requiring researchers to navigate vast literature, formulate hypotheses, design experiments, and synthesize findings \cite{hope2023computational}. The artificial intelligence offers opportunities to transform this landscape \cite{wang2023scientific}.

Among AI techniques, Large Language Models (LLMs) show remarkable capabilities in natural language processing and beyond \cite{naveed2025comprehensive}. The advances demonstrate expanding capabilities in scientific research, including solving challenging mathematical problems \cite{trinh2024solving}, retrieving related work \cite{wang2023scimon}, proposing research ideas \cite{wang2023scimon,yang2024large,baek2025researchagent}, and designing experiments \cite{du2024llms}. More recently, LLM-based multi-agent systems have emerged as a promising approach that specializes LLMs into distinct agents with different capabilities and enables collaboration among them \cite{guo2024large}. The development catalyzes a paradigm shift from AI-assisted research toward automated scientific discovery. Given a topic description, work in this direction integrates the full research pipeline, including literature review, experimentation, and paper writing \cite{lu2024ai,schmidgall2025agent}.

\begin{figure}
\begin{center}
\includegraphics[width=\linewidth]{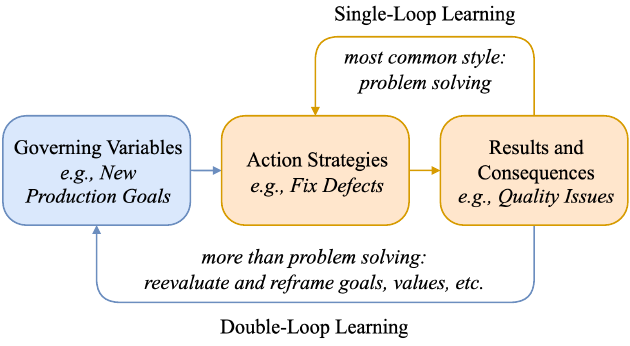}
\end{center}
\caption{Double-loop learning involves two loops. In the first loop, the system takes actions based on the current goals, while in the second loop, the goals themselves can be questioned and modified.}
\label{fig:paradigm}
\end{figure}

Building on these advances, we explore automated research aimed at solving research problems. The challenges of solving research problems can be divided into two folds: First, doing the right things, which means proposing a plan that can effectively solve the research problem. The plan should be technically sound and demonstrate meaningful improvements over existing work. Second, doing things right, which means executing the plan and ensuring that each action step is correctly carried out so that the output of each action meets expectations. For example, claims must be properly supported, mathematical formulations should be error-free, and unexpected experimental results need to be handled appropriately. They define a bilevel optimization problem \cite{colson2007overview}, which is an optimization framework with two hierarchical levels. The upper-level problem seeks promising plans, while the lower-level problem ensures that each action step can be executed correctly.

Can we organize multi-agents to solve the bilevel optimization problem in research? We draw inspiration from the behavioral theory of the firm \cite{cyert1963behavioral}, where organizations make decisions in uncertain environments, and introduce double-loop learning into agent organizations for solving research problems \cite{cyert2015behavioral}. As shown in Fig. \ref{fig:paradigm}, unlike single-loop learning, which operates within one loop and corrects errors to meet current goals, double-loop learning \cite{argyris1977double} involves two loops. In the first loop, the system takes actions based on current goals, while in the second loop, the goals themselves can be questioned and modified. As an example \cite{bratianu2015organizational}, when Toyota faced quality issues in the 1950s, it didn't just fix defects in the first loop, but questioned its entire production philosophy in the second loop, leading to the development of the Toyota Production System.

We propose the Double-Loop Multi-agent Framework involving two loops: the leader loop and the follower loop. The leader loop involves a set of professor agents that aims to optimize plans to solve the research problem. It creates a pool of candidate proposals, which are initialized and optimized in parallel through meetings to explore better plans. The follower loop involves a set of doctoral student agents that aims to take actions correctly and ensure the output of each action meets the plan. Before an action, a meeting is called based on relevant resources to decide the current action, and after the action, the results are used to determine subsequent actions. Finally, a research paper is output to solve the research problem.

% The contributions of our framework are threefold. (1) We introduce the behavioral theory of the firm into multi-agent collaboration, proposing the Double-Loop Multi-agent Framework and opening new avenues for integrating management principles into automatic scientific research. (2) We are the first to propose applying evolutionary algorithms for proposing plans for research problems in the leader loop through involvement meetings, improvement meetings, and integration meetings to explore the solution space. (3) We are the first to explore a dynamic planning mechanism for executing plans in the follower loop, where the current action is dynamically determined by observations, and subsequent actions are determined based on the outputs of the current action to operate under uncertainty.

\section{Related Work}

Large Language Models (LLMs) demonstrate capabilities in natural language processing. The advances demonstrate expanding capabilities in scientific research. For literature review, AutoSurvey \cite{wang2024autosurvey} employs LLMs to automatically write survey papers through iterative retrieval, outline generation, and subsection drafting. For idea generation, SciMON \cite{wang2023scimon} retrieves inspirations from past papers and iteratively optimizes generated ideas through comparison. Si et al. \cite{si2024can} propose an Idea Generation Agent with paper retrieval, generation, and ranking components to produce ideas comparable to human experts. IdeaSynth \cite{pu2025ideasynth} and IRIS \cite{garikaparthi2025iris} enable researchers to iteratively refine ideas through human-computer collaboration. For automatic experimentation, ResearchCodeAgent \cite{gandhi2025researchcodeagent} auto-generates code from existing research papers for benchmarking or building upon literature methods with partial or complete starter code.

\begin{figure*}[h]
\begin{center}
\includegraphics[width=\linewidth]{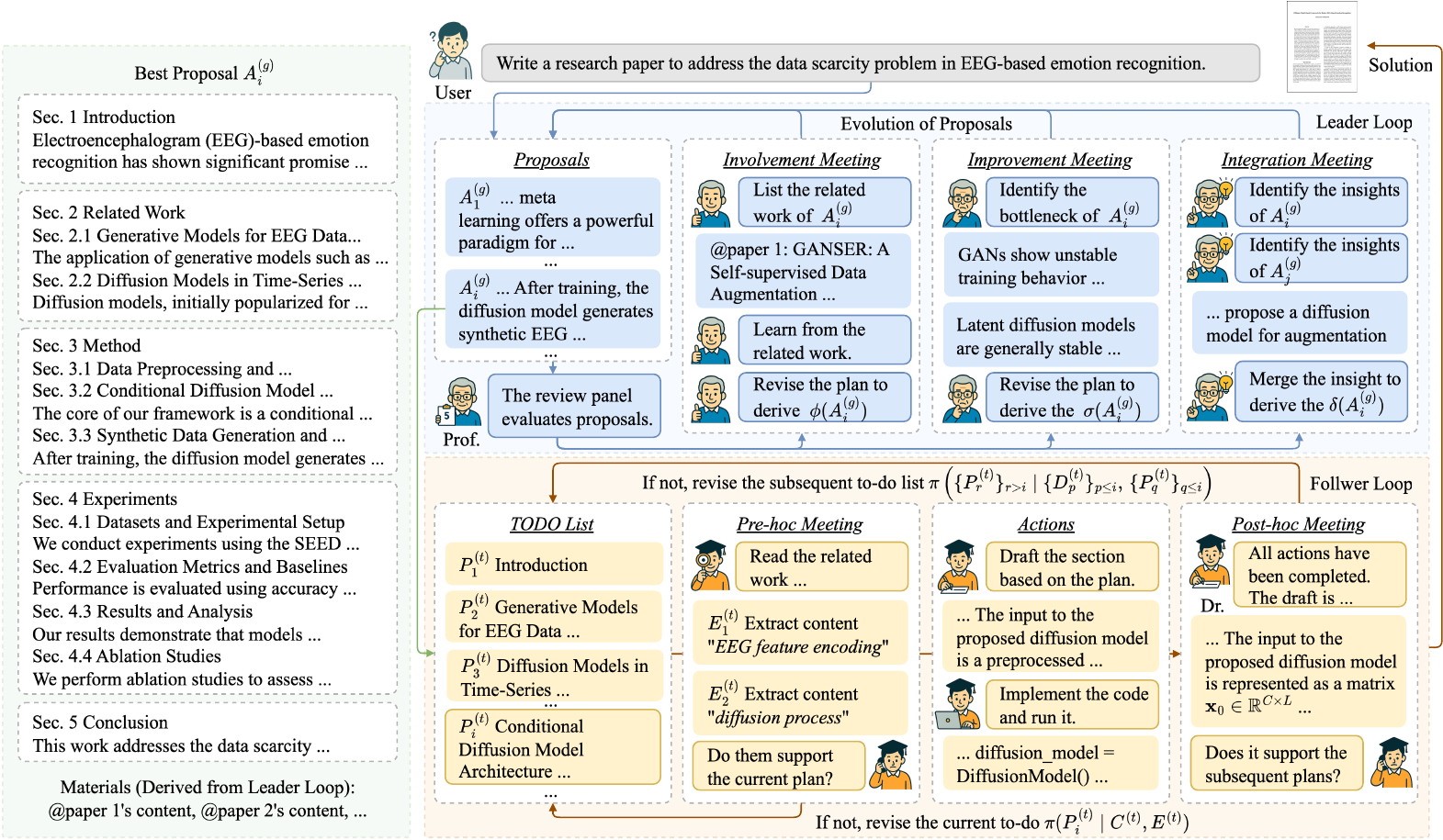}
\end{center}
\caption{Overview of the double-loop learning framework. It consists of two loops: the leader loop and the follower loop. In the leader loop, we build a pool of potential solutions and evolve them iteratively. In the follower loop, we execute the most promising plan and dynamically adapt actions.}
\label{fig:framework}
\end{figure*}

LLM-based multi-agent systems catalyze a paradigm shift from AI-assisted research toward automated scientific discovery. The AI Scientist and AI Scientist-v2 projects \cite{lu2024ai,yamada2025ai} generate research ideas, write code, execute experiments, visualize results, and write scientific papers. Baek et al. propose ResearchAgent \cite{baek2025researchagent} to propose methods and design experiments while iteratively refining them based on feedback from reviewing agents. Weng et al. \cite{weng2025cycleresearcher} introduce an iterative preference training framework with CycleResearcher, which conducts research tasks, and CycleReviewer, which simulates the peer review process and provides iterative feedback via reinforcement learning. However, limited work explores multi-agent frameworks for solving research problems.

\section{Methodology}

\subsection{Problem Formulation}

We formulate automated scientific research as a bilevel optimization problem. Mathematically, it can be denoted as:
\begin{equation}
\begin{aligned}
\max_{p \in \mathcal{P}} \; & R(p, y^*(p)) \\
\text{s.t.} \quad & y^*(p) \in \arg\max_{y \in \mathcal{Y}(p)} f(p, y)
\end{aligned}
\end{equation}
where $p$ represents the upper-level decision variable, the research plan. The feasible set of research plans is denoted as $\mathcal{P}$, and $R$ is the upper-level objective function representing the rating from a review panel. Meanwhile, $y$ represents the lower-level decision variable, i.e., the action sequence for executing the plan. The feasible set of action sequences is denoted by $\mathcal{Y}(p)$, and $f$ is the lower-level objective function that measures whether action inputs and outputs conflict with $p$.

As shown in Fig. \ref{fig:framework}, we propose the double-loop multi-agent (DLMA) framework for the bilevel optimization. Given a user input $U$ describing a research problem, the DLMA framework aims to generate a research paper that addresses the problem. It consists of two loops, the leader loop and the follower loop, which we describe below.

\subsection{Leader Loop}

The leader loop involves a set of professor agents. They take a research problem provided by the user as input and output a plan to optimize $\max_{p \in \mathcal{P}} R(p, y^*(p))$. The professor agents initialize a population of $N$ potential proposals as $\mathcal{T}^{(0)} = \{t_1^{(0)}, t_2^{(0)}, \ldots, t_N^{(0)}\}$. The leader loop follows an iterative paradigm in which the population $\mathcal{T}^{(g)}$ updates via three types of meetings: involvement meeting $\phi(\cdot)$, which introduces proposals from the perspectives of references; improvement meeting $\sigma(\cdot)$, which refines existing proposals by addressing their weaknesses; and integration meeting $\delta(\cdot)$, which combines the strengths from different proposals.

The involvement meeting introduces existing research solutions to enrich the proposal pool with diverse perspectives on the research problem. Given the user request $U$, we prompt the LLM to formulate a query and search for relevant papers\footnote{https://huggingface.co/docs/hub/api}. Referring to the retrieved references, we prompt the LLM to generate a potential proposal. Mathematically, the derived result can be denoted as $\phi(U)$, where $\phi(\cdot)$ represents the involvement meeting. In the initial generation ($g=0$), we employ only the involvement meeting to populate the solution space.

The improvement meeting samples a proposal $t_i^{(g)} \in \mathcal{T}^{(g)}$ and performs a critical reflection process that identifies weaknesses. The improvement meeting agent creates queries to search for information to address the identified weaknesses. Then, we prompt the LLM to revise proposal $t_i^{(g)}$ to address its weaknesses. Mathematically, the derived result can be denoted as $\sigma(t_i^{(g)})$, where $\sigma(\cdot)$ represents the improvement meeting.

The integration meeting samples a pair of proposals $(t_i^{(g)}, t_j^{(g)}) \in \mathcal{T}^{(g)} \times \mathcal{T}^{(g)}$, where $i \neq j$. The professor agents analyze the strengths of each proposal. We then execute a prompting process to generate new proposals that integrate complementary strengths. This produces two offspring: $\delta(t_i^{(g)}, t_j^{(g)})$, which integrates strengths from $t_j^{(g)}$ into $t_i^{(g)}$, and $\delta(t_j^{(g)}, t_i^{(g)})$, which integrates strengths from $t_i^{(g)}$ into $t_j^{(g)}$. Mathematically, the derived results can be denoted as $\mathcal{T}_{\delta}^{(g)}$, where $\delta(\cdot)$ represents the integration meeting.

After the involvement meeting, improvement meeting, and integration meeting produce new proposals $\mathcal{T}_{\phi}^{(g)}$, $\mathcal{T}_{\sigma}^{(g)}$, and $\mathcal{T}_{\delta}^{(g)}$, they are combined with the original population $\mathcal{T}^{(g)}$. To guide the evolutionary process toward optimal proposals, we design a review panel that evaluates proposals. We collect review forms from OpenReview\footnote{https://openreview.net/} and prompt the LLM to evaluate each proposal and provide feedback comments based on which a rating can be derived, denoted by $\mu(\cdot)$. Then, we implement a selection mechanism to select the top $K$ proposals with the highest rating scores. Mathematically, this can be denoted as
\begin{equation}
\mathcal{T}^{(g+1)} = \epsilon\left(\mathcal{T}^{(g)} \cup \mathcal{T}_{\phi}^{(g)} \cup \mathcal{T}_{\sigma}^{(g)} \cup \mathcal{T}_{\delta}^{(g)}\right)
\end{equation}
where $\epsilon(\cdot)$ is the selection operation based on the scores of proposals by $\mu(\cdot)$. The selection process continues iteratively until either the defined number of evolution cycles is completed or the convergence criteria are met, i.e., when the improvement in the maximum reward falls below a threshold $\varepsilon$. The final output is the highest-rating proposal:
\begin{equation}
P = \arg\max_{t \in \mathcal{T}^{(G)}} \mu(t)
\end{equation}
where $G$ is the final generation.

\subsection{Follower Loop}

The best proposal $P$ is then passed to the follower loop, which is responsible for executing $P$. The follower loop involves a set of doctoral student agents that aim to optimize $\arg\max_{y \in \mathcal{Y}(p)} f(p, y)$. We start with $P^{(0)} = P$. We format the best proposal as a to-do list to draft a research paper. Here, we denote $s \in \{1, \ldots, S\}$, where $S$ is the total number of steps. Then, the to-do list is updated whenever discrepancies between the proposal and the action input/output arise. We denote the rounds of updates as $t$, and the to-do list at round $t$ is denoted as $P^{(t)} = \{P_1^{(t)}, \ldots, P_S^{(t)}\}$.

In detail, before taking action, we first call a pre-hoc meeting for doctoral agents. When processing step $s$ in round $t$, we first fetch observations relevant to $P_s^{(t)}$. We focus on two types of observations: contextual observations fetched from the existing draft, and external observations fetched from reference papers. The LLM is prompted to locate relevant content, identifying appropriate line ranges from both external references and the existing draft. The content retrieved from the existing draft is denoted as $C_s$, while the content retrieved from external references for step $s$ is denoted as $E_s$. Based on these observations, we update the current step of the to-do list:
\begin{equation}
P_s^{(t+1)} = \pi(P_s^{(t)} \mid C_s, E_s)
\end{equation}
where $\pi(\cdot)$ is a policy function that revises the $s$-th step based on the contextual observation $C_s$ and external observation $E_s$. The policy function is implemented by prompting the LLM to revise the current step of the to-do list.

After the pre-hoc meeting, the doctoral agents take actions. The revised proposal $P_s^{(t+1)}$ is then executed to draft $D_s$ for the $s$-th section, with the contextual observation $C_s$, external observation $E_s$, and $P_s^{(t+1)}$. To ensure the draft has correct LaTeX syntax, we render the draft by compiling it in a LaTeX project to obtain error logs and warning logs. Based on the logs, we prompt the LLM to revise the drafted manuscript and recompile it, until there are no errors or warnings, or until a maximum number of revisions is reached.

In addition to drafting, we maintain a code project during execution, which is updated at each step if the LLM determines it is necessary. We follow PaperBench and implement the IterativeAgent \cite{staracepaperbench}. We provide the agent with a bash shell command execution tool, a Python code execution tool, a web browser tool, and a paginated file reader tool. The agent operates a tool-use loop until the model chooses to terminate its run or a time limit is reached. After termination, the code project and execution logs are integrated via Repomix\footnote{https://repomix.com/}, serving to update the draft.

After the action is complete, we call a post-hoc meeting. After generating the draft $D_s$, we update the subsequent steps in the to-do list to ensure consistency with the newly derived content. For each subsequent step $q > s$, the to-do list is revised as:
\begin{equation}
P_q^{(t+1)} = \pi\left(P_q^{(t)} \mid D_s, \{D_p\}_{p < s}, \{P_r^{(t+1)}\}_{r < q}\right)
\end{equation}
where the policy function $\pi(\cdot)$ revises each subsequent step $P_q^{(t)}$ by considering the current section's draft $D_s$, previous section drafts $\{D_p\}_{p < s}$, and previous to-do steps $\{P_r^{(t+1)}\}_{r < q}$ as context. The LLM is prompted to revise these subsequent steps if they are no longer suitable; otherwise, the original steps are retained. For steps that have already been executed ($q < s$), the to-do list remains unchanged:
\begin{equation}
P_q^{(t+1)} = P_q^{(t)}
\end{equation}

Next, we proceed to step $s+1$ and begin a new round of iteration. This process continues until all $S$ steps are completed. We collect the final drafts, $\{D_1, D_2, \ldots, D_S\}$, and then render the LaTeX project, and the final PDF of the paper is generated.

\section{Experiments}

\begin{table}
\centering
\begin{adjustbox}{width=\columnwidth}
\begin{tabular}{lcccc}
\toprule
Method & Soundness $\uparrow$ & Excitement $\uparrow$ & Overall $\uparrow$ & Confidence $\uparrow$ \\
\midrule
GPT-5 & 3.68 & 3.77 & 3.78 & 3.95 \\
Gemini 2.5 Pro & 3.78 & 3.85 & 3.83 & 4.02 \\
Claude Sonnet 4 & 3.75 & 3.82 & 3.81 & 4.01 \\
CycleResearcher & 3.77 & 3.84 & 3.82 & 4.02 \\
Agent Laboratory & 3.92 & 3.93 & 3.94 & 4.08 \\
Dolphin & 3.88 & 3.95 & 3.93 & 4.07 \\
\midrule
\cellcolor{gray!15}Proposed & \cellcolor{gray!15}\textbf{4.03} & \cellcolor{gray!15}\textbf{4.02} & \cellcolor{gray!15}\textbf{4.01} & \cellcolor{gray!15}\textbf{4.13} \\
\bottomrule
\end{tabular}
\end{adjustbox}
\caption{Performance comparison on the ACLAward dataset.}
\label{tab:comaprative_experiment_1}
\end{table}

\begin{table}
\centering
\begin{adjustbox}{width=\columnwidth}
\begin{tabular}{lcccc}
\toprule
Method & Soundness $\uparrow$ & Excitement $\uparrow$ & Overall $\uparrow$ & Confidence $\uparrow$ \\
\midrule
GPT-5            & 3.42 & 3.08 & 3.09 & 3.98 \\
Gemini 2.5 Pro   & 3.51 & 3.18 & 3.17 & 4.05 \\
Claude Sonnet 4  & 3.48 & 3.15 & 3.14 & 4.03 \\
CycleResearcher  & 3.50 & 3.17 & 3.16 & 4.04 \\
Agent Laboratory & 3.87 & 3.42 & 3.45 & 4.35 \\
Dolphin          & 3.66 & 3.38 & 3.43 & 4.34 \\
\midrule
\cellcolor{gray!15}Proposed & \cellcolor{gray!15}\textbf{3.93} & \cellcolor{gray!15}\textbf{3.68} & \cellcolor{gray!15}\textbf{3.69} & \cellcolor{gray!15}\textbf{4.68} \\
\bottomrule
\end{tabular}
\end{adjustbox}
\caption{Performance comparison on the Laboratory dataset.}
\label{tab:agent_laboratory_comparison}
\end{table}

In this section, we conduct experiments on three benchmark datasets.

\textbf{ACLAward}: To establish the human expert baseline, we collect award-winning papers from The 63rd Annual Meeting of the Association for Computational Linguistics (ACL 2025). We annotate 10 well-defined research problems from award papers. For each research problem, we generate 10 papers using our algorithm, resulting in 100 research papers in total.

\begin{figure*}
\begin{center}
\includegraphics[width=\linewidth]{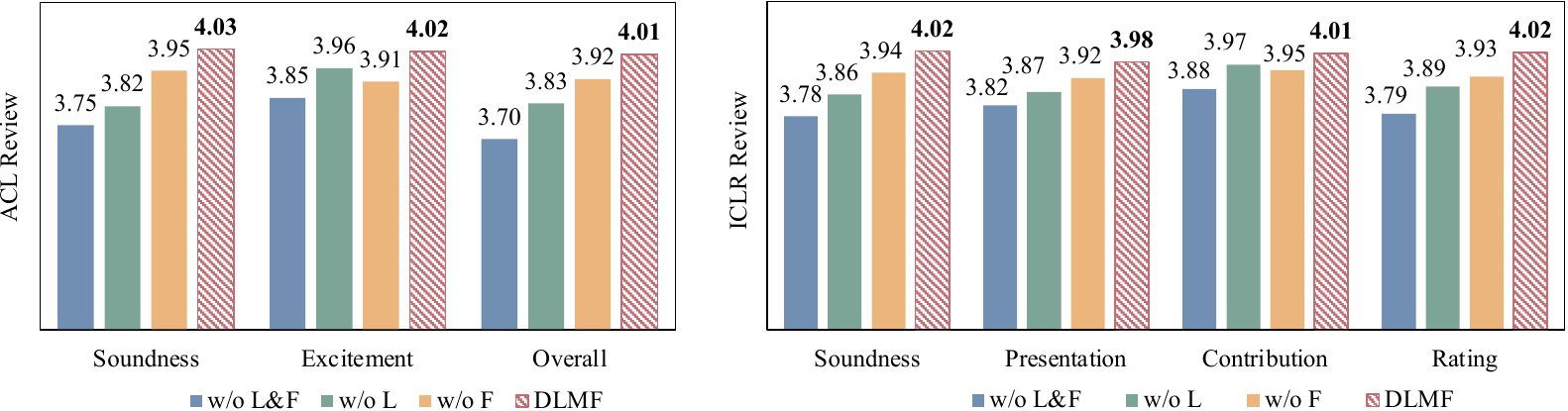}
\end{center}
\caption{Ablation study on the ACLAward dataset with the ACL review and ICLR review.}
\label{fig:ablation_study}
\end{figure*}

\textbf{Laboratory}: Recent studies have explored autonomous LLM-based frameworks capable of completing the entire research process. To compare with state-of-the-art algorithms, we follow \cite{schmidgall2025agent} to collect five research problems. These topics cover natural language processing and computer vision. We generate 10 papers using our algorithm for each problem, resulting in 50 research papers in total.

\textbf{Plagiarism}: Recent studies have shown that a considerable fraction of automatically generated research documents are smartly plagiarized. We follow~\cite{gupta2025all} to derive 12 research problems in natural language processing. Based on these research problems, we generate 10 papers for each problem, resulting in 120 research papers in total.

\subsection{Comparative Experiments}
\label{sec:comparative_experiments}

We first conduct comparative experiments on the ACLAward and Laboratory datasets. We use GPT-5 as the backbone model to construct agent collaboration. We set the temperature to 0.3. For hyperparameters, we set the number of generations in the leader loop to 5 and the proposal number to 10. In each generation, we set the probability to call meetings to update proposals: involvement meeting to 10\%, improvement meeting to 30\%, integration meeting to 50\%, and 10\% remain unchanged.

To evaluate paper quality, we assess generated papers using ACL-style criteria following the ACL review form\footnote{https://aclrollingreview.org/reviewform}. We adopt the LLM-as-a-judge for reproducible automatic evaluation, following \cite{weng2025cycleresearcher,schmidgall2025agent}. The LLM is prompted to generate feedback including Paper Summary, Summary of Strengths, Summary of Weaknesses, and Comments/Suggestions/Typos. It then provides ratings for Reviewer Confidence (1-5), Soundness (1-5), Excitement (1-5), and Overall Assessment (1-5).

As shown in Table~\ref{tab:comaprative_experiment_1}, we compare the proposed method with state-of-the-art large language models, including GPT-5, Gemini 2.5 Pro, and Claude Sonnet 4. We also compare with multi-agent frameworks for automated scientific research, including CycleResearcher~\cite{weng2025cycleresearcher}, Agent Laboratory~\cite{schmidgall2025agent}, and Dolphin~\cite{yuan2025dolphin}. We observe that vanilla large language models directly generating research papers still have substantial room for improvement to be directly accepted by ACL, compared with human experts who are able to produce award-winning papers. This indicates the difficulty of scientific research. CycleResearcher achieves scores comparable to but slightly outperforming vanilla large language models, benefiting from data-driven fine-tuning on research papers. Recent multi-agent collaboration frameworks achieve a breakthrough with average scores exceeding 3.9, demonstrating their capability in handling complex tasks with tools. The proposed method further achieves the best performance. In particular, the performance gap on soundness indicates that the proposed method better finds reasonable solutions to solve research problems.

To verify the consistency between LLM evaluation methods and human evaluation, we follow \cite{wang2024autosurvey} to conduct a meta-evaluation. We sample 20 generated papers, 2 for each research problem. Then, human experts judge pairs of generated papers to determine which one is superior. We provide the experts with the same scoring criteria used in our evaluation for reference. Finally, the experts rank the generated 20 papers. We compare these rankings with those generated by LLMs using Spearman's rank correlation coefficient. The LLM evaluation achieves a correlation of 0.4609, suggesting LLMs align well with human judgment and provide a reliable proxy, though we observe human ratings have greater variance indicating stronger discrimination.

We then conduct comparative experiments on the Laboratory dataset following the same settings. As shown in Table~\ref{tab:agent_laboratory_comparison}, we observe lower scores for both soundness and excitement, leading to lower overall scores. This suggests that research problems affect model outputs. The ACLAward dataset contains novel and impactful research problems from award-winning papers, which leads to higher ratings for generated papers. Compared with the ACLAward dataset, the Laboratory dataset simulates the model's performance in handling research questions that are less rigorous and engaging rather than problems from award-winning papers. Even so, the proposed method outperforms other algorithms in generating scientific papers.

\subsection{Ablation Study}

\begin{figure*}
\begin{center}
\includegraphics[width=\linewidth]{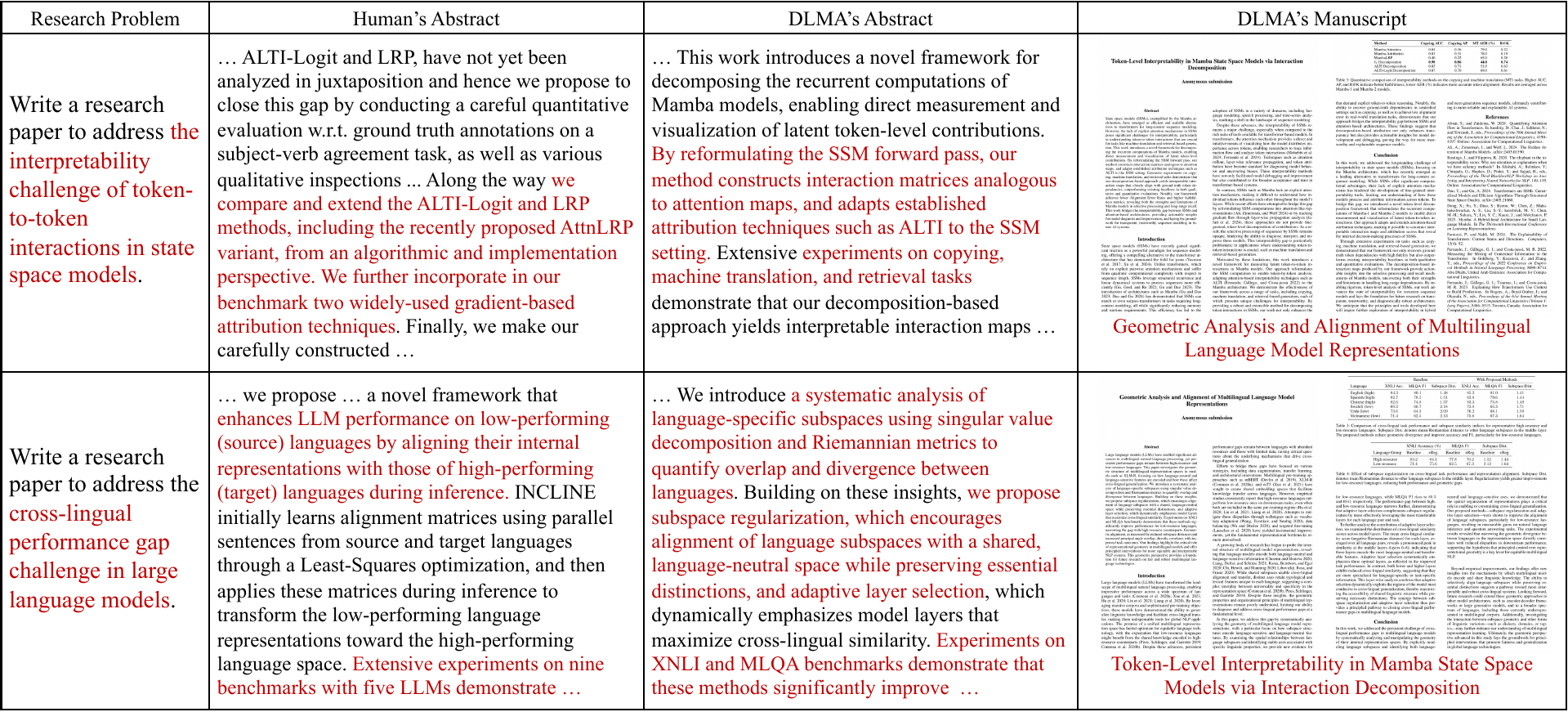}
\end{center}
\caption{Two case studies comparing human expert research output and Double-Loop Multi-agent (DLMA) Framework output.}
\label{fig:case_study}
\end{figure*}

In this section, we conduct ablation studies to evaluate the performance gain from the proposed modules. When ablating the evolution module, we do not perform involvement, improvement, and integration meeting operations, instead initializing with a single plan generated by the LLM, denoted as Adaptation. When ablating the adaptation module, we use a fixed plan instead. When errors arise, we attempt to fix them without replanning and, after reaching the maximum number of attempts, skip to the following section. We denote this version as Evolution.

To evaluate paper quality, we assess generated papers on the ACLAward dataset using different review criteria to avoid bias. Besides ACL, following \cite{weng2025cycleresearcher}, we adopt the review form from the International Conference on Learning Representations (ICLR). The LLM first generates qualitative feedback including Summary, Strengths, Weaknesses, and Questions. Based on this feedback, it then provides ratings for Soundness (1-4), Presentation (1-4), Contribution (1-4), Rating (1-10), and Confidence (1-5). We normalize them to 1 to 5 to align with ACL review results.

As shown in Fig. \ref{fig:ablation_study}, ablating either evolution or adaptation leads to performance degradation regardless of the review criteria, demonstrating the effectiveness of the proposed modules. We observe that removing adaptation causes the largest decrease in Soundness, indicating that adaptation helps align observations with plans and ensures solid technical details. Meanwhile, evolution contributes most to Excitement and Contribution. This shows that the leader loop explores more novel solutions and generates more exciting and contributive proposals.

\subsection{Case Studies}

In this section, we conduct case studies on two research problems sampled from ACL award-winning papers. We compare the outputs of human experts and the Double-Loop Multi-agent (DLMA) framework. As shown in Fig. \ref{fig:case_study}, the first row demonstrates the human expert's ACL award paper \cite{wang2024bridging} that addresses the interpretability challenge of token-to-token interactions in state space models. The second row demonstrates the human expert's ACL award paper \cite{pitorro2025latim} that addresses the cross-lingual performance gap challenge in large language models. We present the abstracts from both the original papers and DLMA-generated papers, and provide the generated manuscripts for demonstration.

In the first case, the research problem is novel and lacks existing work. We find that both the DLMA-generated output and the human expert's work employ attribution techniques, specifically ALTI, for state space model interpretability. This shows that the proposed method can identify appropriate techniques at a human level. However, the human expert further incorporates advanced techniques such as LPR and AAttnLRP. The generated method lacks such further improvements due to limitations in coding ability for complex projects. In the second case, the research problem has been studied before, but both the human expert and DLMA propose new perspectives to address the problem.

\subsection{Discussions on Doing Right Things}

In this section, we conduct experiments to gain deeper insights into the designed leader loop. We conduct experiments on the Plagiarism dataset following the experimental settings in Section~\ref{sec:comparative_experiments} and then evaluate the proposals for each generation. We record the minimum, mean, and maximum evaluation scores of the plans in each generation. To avoid bias and randomness in ACL review results, we also record evaluation results using the NeurIPS review form. Following \cite{schmidgall2025agent}, we adopt the review form from the Conference on Neural Information Processing Systems (NeurIPS). The LLM first generates qualitative feedback including Summary, Strengths and Weaknesses, Questions, and Limitations. Based on this feedback, it then provides ratings for Quality (1-4), Clarity (1-4), Significance (1-4), Originality (1-4), Overall (1-6), and Confidence (1-5). The results are reported in Table~\ref{tab:generation_evolution}.

We observe that regardless of review criteria and review form, the mean evaluation score improves as the generation increases. This demonstrates that the proposed method can find better solutions to research problems during evolution. Meanwhile, we observe that the improvement gradually plateaus after the 4th generation, with the mean ACL score in the 5th generation remaining the same as in the 4th generation. This suggests that there exists a bottleneck for LLM-based automated scientific research to achieve unlimited self-improvement.

\subsection{Discussions on Doing Things Right}

\begin{table}
\centering
\begin{adjustbox}{width=\columnwidth}
\begin{tabular}{ccccccc}
\toprule
\multirow{2}{*}{Generation} & \multicolumn{3}{c}{ACL Review Form} & \multicolumn{3}{c}{NeurIPS Review Form} \\
\cmidrule(lr){2-4} \cmidrule(lr){5-7}
 & Min $\uparrow$ & Mean $\uparrow$ & Max $\uparrow$ & Min $\uparrow$ & Mean $\uparrow$ & Max $\uparrow$ \\
\midrule
1 & 3.00 & 3.86 & 4.00 & 3.00 & 4.04 & 5.00 \\
2 & 3.50 & 3.93 & 4.00 & 3.00 & 4.06 & 5.00 \\
3 & 3.50 & 3.98 & 4.00 & 4.00 & 4.07 & 5.00 \\
4 & 3.50 & 3.99 & 5.00 & 4.00 & 4.11 & 5.00 \\
5 & 3.50 & 3.99 & 5.00 & 4.00 & 4.15 & 5.00 \\
\bottomrule
\end{tabular}
\end{adjustbox}
\caption{ACL and NeurIPS review form on the proposals across generations on the Plagiarism dataset.}
\label{tab:generation_evolution}
\end{table}

In this section, we conduct experiments to gain deeper insights into the designed follower loop. We conduct experiments on the ACLAward dataset. We collect observations and to-do lists before the pre-hoc meeting and after the post-hoc meeting. We then prompt the LLM to check whether the observations can support the to-do list and record the result. Next, we compute the support rate for each follower loop before the pre-hoc meeting and after the post-hoc meeting.

\begin{figure}
\begin{center}
\includegraphics[width=\linewidth]{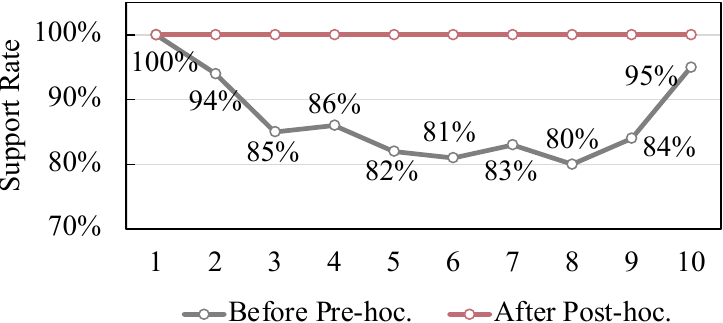}
\end{center}
\caption{Support rate of the $t$-th planning in the to-do list before the pre-hoc meeting and after the post-hoc meeting. Higher scores indicate better alignment between observations and the plan.}
\label{fig:support_rate}
\end{figure}

We report the support rate of the first 10 steps in Fig.~\ref{fig:support_rate}. We observe that in the beginning steps of the follower loop, the plan before the pre-hoc meeting is well supported even without replanning, showing minimal knowledge gaps in the introduction section of research papers. However, as the steps progress, the support rate decreases. This decrease stems from issues such as failure to retrieve relevant references, mismatches between the selected algorithm and the research problem, and experimental results that deviate from expectations. In contrast, after the post-hoc meeting, the support rate remains high, showing the follower loop aligns planning and observations.

\section{Limitations}

In this section, we discuss the limitations of our work. Although the proposed method improves the automated scientific research, the leader loop and follower loop incur significant computational costs compared to trivial pipelines, due to searching the solution space and dynamically adjusting plans. We record the average time cost of different ablated versions of the proposed method without including experimental runtime as it is research problem-specific, and report the results in Table~\ref{tab:computational_cost}. We observe that almost half an hour and two million tokens are typically required. Besides, we find that the current code agent can hallucinate, leading to implementations that differ from their descriptions. In future work, we plan to focus on paper and code alignment to make experimental results more reliable.

\begin{table}
\centering
\begin{adjustbox}{width=0.6\columnwidth}
\begin{tabular}{cccc}
\toprule
Leader & Follower & Time (s) $\downarrow$ & Tokens $\downarrow$ \\
\midrule
\xmark & \xmark & \textbf{212} & \textbf{575,005} \\
\cmark & \xmark & 1,016 & 859,461 \\
\xmark & \cmark & 378 & 787,136 \\
\cellcolor{gray!15}\cmark & \cellcolor{gray!15}\cmark & \cellcolor{gray!15}1,558 & \cellcolor{gray!15}1,751,901 \\
\bottomrule
\end{tabular}
\end{adjustbox}
\caption{Computational cost comparison across ablated versions.}
\label{tab:computational_cost}
\end{table}

\section{Conclusion}

In this work, we introduce the Double-Loop Multi-Agent (DLMA) framework to tackle the automated scientific research problem by explicitly decomposing it into two interdependent challenges: evolving high-quality research plans and executing them reliably. The leader loop addresses the first challenge by evolving a population of proposals through evolutionary meetings, effectively exploring the solution space for novel and promising plans. The follower loop addresses the second challenge by dynamically executing the selected plan, ensuring each step is aligned with observations and correctly implemented. Extensive evaluations demonstrate that our framework generates research papers with superior quality, outperforming existing state-of-the-art methods. Ablation studies further verify that the evolution mechanism is crucial for fostering excitement and contribution, while the execution mechanism is vital for ensuring soundness and technical solidity.

% \bibliographystyle{ACM-Reference-Format}

\bibliography{custom}

\end{document}